\documentclass[11pt]{article} 

\usepackage{fullpage}

\usepackage{comment}
\usepackage{url}
\usepackage{amsmath}
\usepackage{graphicx,xspace}

\usepackage{microtype}
\usepackage{booktabs} 
\usepackage{amsfonts}       
\usepackage{nicefrac}       
\usepackage{microtype}      

\usepackage{caption}
\usepackage{subcaption}
\usepackage{color}
\usepackage{mdwlist}

\usepackage{paralist}
\usepackage{algorithm,algorithmic}
\newfloat{Algorithm}{tb}{lox}
\floatname{Algorithm}{Algorithm}

\usepackage[colorlinks,linkcolor=red,citecolor=blue,urlcolor=blue]{hyperref}
\usepackage{caption}
\usepackage{subcaption}
\DeclareCaptionType{copyrightbox}
\usepackage{tabularx}
\usepackage{etoolbox}
\usepackage{mdwlist}
\usepackage{breakurl}

\usepackage[numbers,sort&compress]{natbib}

\usepackage{graphicx} 
\usepackage{tikz}
\usetikzlibrary{positioning,patterns,shapes, shadows}
\usepackage{pgfplots}
\usepgfplotslibrary{external}
\usepackage{multirow}
\usepackage{rotating}
\usepackage{wrapfig}
\usepackage{array} 
\usepackage{pbox}

\colorlet{darkgreen}{green!50!black}
\colorlet{darkred}{red!90!black}
\colorlet{darkyellow}{yellow!90!black}
\colorlet{darkgreen}{green!50!black}
\colorlet{darkred}{red!90!black}
\colorlet{darkyellow}{yellow!90!black}

\definecolor{Green}{rgb}{0.0, 0.5, 0.0}

\definecolor{blue-green}{rgb}{0.0, 0.87, 0.87}
\definecolor{brightturquoise}{rgb}{0.03, 0.91, 0.87}
\definecolor{darkturquoise}{rgb}{0.0, 0.81, 0.82}

\tikzset{
  gnode/.style={
    fill=white,
    draw=black,
    circle,
    very thick, 
    inner sep=3.5,
    drop shadow={shadow xshift=0.3ex,shadow yshift=-0.5ex, path
      fading={circle with fuzzy edge 20 percent}}
  }
}

\tikzset{
  rnode/.style={
    fill=black,
    draw=black,
    circle,
    very thick, 
    inner sep=3.5,
    drop shadow={shadow xshift=0.3ex,shadow yshift=-0.5ex, path
      fading={circle with fuzzy edge 20 percent}}
  }
}

\tikzset{
  ynode/.style={
    fill=black!50!white,
    draw=black,
    circle,
    very thick, 
    inner sep=3.5,
    drop shadow={shadow xshift=0.3ex,shadow yshift=-0.5ex, path
      fading={circle with fuzzy edge 20 percent}}
  }
}

\usepackage{amssymb}
\usepackage{pifont}
\usepackage{natbib}

\usepackage{algorithm}
\usepackage{algorithmic}

\input{./Definitions}
\graphicspath {{./Figures/}}

\begin{document}

\def\gridwidth{2}
  \def\nodenum{4}
  \def\asnwidth{0.5}
  \def\asnmargin{0.05}

\newcommand{\asnbox}[3]{
  \fill [#3!25!lightgray]
  (#1 * \asnwidth + \asnmargin, 
  \gridwidth * 3 - #2 * \gridwidth + \asnmargin)
  rectangle
  (#1 * \asnwidth + \asnwidth - \asnmargin, 
  \gridwidth * 3 - #2 * \gridwidth + \gridwidth - \asnmargin);
  \draw
  (#1 * \asnwidth + \asnmargin, 
  \gridwidth * 3 - #2 * \gridwidth + \asnmargin)
  rectangle
  (#1 * \asnwidth + \asnwidth - \asnmargin, 
  \gridwidth * 3 - #2 * \gridwidth + \gridwidth - \asnmargin);
}

  \newcommand{\dasnbox}[2]{
    \draw[dashed]
    (#1 * \asnwidth + \asnmargin, 
    \gridwidth * 3 - #2 * \gridwidth + \asnmargin)
    rectangle
    (#1 * \asnwidth + \asnwidth - \asnmargin, 
    \gridwidth * 3 - #2 * \gridwidth + \gridwidth - \asnmargin);
  }

  \newcommand{\dtpt}[2]{
    \draw (#1 * \asnwidth + 0.5 * \asnwidth, 
    #2 * \asnwidth + 0.5 * \asnwidth) 
    node {\tiny{x}};
  }

  \newcommand{\mvarr}[4]{
    \draw [->]
    (#1 * \asnwidth + 0.5 * \asnwidth, 
    3 * \gridwidth - #2 * \gridwidth + 0.5 * \gridwidth) 
    --
    (#3 * \asnwidth + 0.5 * \asnwidth, 
    3 * \gridwidth - #4 * \gridwidth + 0.5 * \gridwidth);
  }
  
  \newcommand{\boxlayout}{
    \draw (0,0) rectangle (\gridwidth * \nodenum, \gridwidth * \nodenum);

    \fill [red!10] (0, 3 * \gridwidth) 
    rectangle (\gridwidth * \nodenum, 4 * \gridwidth);

    \fill [green!10] (0, 2 * \gridwidth) 
    rectangle (\gridwidth * \nodenum, 3 * \gridwidth);

    \fill [blue!10] (0, 1 * \gridwidth) 
    rectangle (\gridwidth * \nodenum, 2 * \gridwidth);

    \fill [brown!10] (0,0) 
    rectangle (\gridwidth * \nodenum, 1 * \gridwidth);

    \foreach \y in {1,2,3}
    {
      \draw[densely dashed] 
      (-\gridwidth * 0.1, \gridwidth * \y)
      --
      (\gridwidth * \nodenum + \gridwidth * 0.1,  \gridwidth * \y);
    }

    \dtpt{5}{1}  \dtpt{0}{15}  \dtpt{13}{12}  \dtpt{13}{6}  \dtpt{14}{8}  \dtpt{0}{8}  \dtpt{7}{10}  \dtpt{15}{2}  \dtpt{10}{5}  \dtpt{8}{13}  \dtpt{5}{3}  \dtpt{12}{5}  \dtpt{10}{7}  \dtpt{11}{6}  \dtpt{15}{3}  \dtpt{6}{9}  \dtpt{13}{5}  \dtpt{1}{3}  \dtpt{8}{15}  \dtpt{1}{6}  \dtpt{1}{7}  \dtpt{8}{9}  \dtpt{12}{4}  \dtpt{12}{1}  \dtpt{13}{10}  \dtpt{9}{0}  \dtpt{8}{4}  \dtpt{15}{12}  \dtpt{3}{6}  \dtpt{10}{13} 
  }

  \newcommand{\ptrna}{crosshatch}
  \newcommand{\ptrnb}{north west lines}
  \newcommand{\ptrnc}{north east lines}
  \newcommand{\ptrnd}{horizontal lines}

  \newcommand{\ptcla}{red}
  \newcommand{\ptclb}{green}
  \newcommand{\ptclc}{blue}
  \newcommand{\ptcld}{brown}

  \newcommand{\drawrowpart}{
    \draw[pattern=\ptrnd, pattern color=\ptcld] 
    (-\gridwidth - \gridwidth * 0.25, 0 * \gridwidth) 
    rectangle 
    (- \gridwidth * 0.25, 
    0 * \gridwidth + 1 * \gridwidth);
    \draw[pattern=\ptrnc, pattern color=\ptclc] 
    (-\gridwidth - \gridwidth * 0.25, 
    1 * \gridwidth) 
    rectangle 
    (- \gridwidth * 0.25, 
    1 * \gridwidth + 1 * \gridwidth);
    
    \draw[pattern=\ptrnb, pattern color=\ptclb] 
    (-\gridwidth - \gridwidth * 0.25, 
    2 * \gridwidth) 
    rectangle 
    (- \gridwidth * 0.25, 
    2 * \gridwidth + 1 * \gridwidth);
    
    \draw[pattern=\ptrna, pattern color=\ptcla] 
    (-\gridwidth - \gridwidth * 0.25, 
    3 * \gridwidth) 
    rectangle 
    (- \gridwidth * 0.25, 
    3 * \gridwidth + 1 * \gridwidth);
  }
  
  \newcommand{\drawcolpart}[4]{
    \draw[pattern=#3, pattern color=#4] 
    (#1 * \asnwidth, 
    4 * \gridwidth + 0.25 * \gridwidth) 
    rectangle 
    (#2 * \asnwidth + \asnwidth, 
    4 * \gridwidth + 0.25 * \gridwidth + \gridwidth);
  }


\title{DS-FACTO: Doubly Separable Factorization Machines}
\author{
Parameswaran Raman\\
       {University of California Santa Cruz}\\
       {params@ucsc.edu}
\and
S.V.N. Vishwanathan\\
        {Amazon}\\
        {vishy@amazon.com}
}
\date{}


\maketitle

\begin{abstract}
Factorization Machines (FM) are powerful class of models that incorporate higher-order interaction among features to add more expressive power to linear models. They have been used successfully in several real-world tasks such as click-prediction, ranking and recommender systems. Despite using a low-rank representation for the pairwise features, the memory overheads of using factorization machines on large-scale real-world datasets can be prohibitively high. For instance on the criteo tera dataset, assuming a modest $128$ dimensional latent representation and $10^{9}$ features, the memory requirement for the model is in the order of $1$ TB. In addition, the data itself occupies $2.1$ TB. Traditional algorithms for FM which work on a single-machine are not equipped to handle this scale and therefore, using a distributed algorithm to parallelize the computation across a cluster is inevitable. In this work, we propose a hybrid-parallel stochastic optimization algorithm DS-FACTO, which partitions both the data as well as parameters of the factorization machine simultaneously. Our solution is fully de-centralized and does not require the use of any parameter servers. We present empirical results to analyze the convergence behavior, predictive power and scalability of DS-FACTO.
\end{abstract}

\section{Introduction}
Factorization Machines (FM), introduced by \citep{Ren10} are powerful class of models which combine the benefits of polynomial regression and computational benefits of low-rank latent variable models such as matrix factorization. They offer a principled and flexible framework to model a variety of machine learning tasks. With the suitable feature representation, they can be used to model tasks ranging from regression, classification, learning to rank, collaborative filtering to temporal models. This makes FM a universal workhorse for predictive modeling as well as building ranking and recommender systems.

Factorization machines present a novel way to represent the higher-order interactions using low-rank latent embeddings of the features. A second-order FM thus requires a model storage of $\Ocal \rbr{K \times D}$, where $K$ is the number of latent dimensions for a feature and $D$ is the number of features. While this is substantially smaller than the dense parameterization of polynomial regression, which would require $\Ocal \rbr{D^{2}}$ storage, we notice that training the FM model is still fraught will computational challenges. For example, consider the Criteo click logs dataset \citep{criteoteralogs14}. Running FM on this dataset even with a modest latent representation of $K=128$ and $10^{9}$ features would easily require memory in the order of $1$ GB for the model parameters. In addition, the data itself occupies $2.1$ TB. Such loads are impossible to run using a single-machine algorithm and demands developing distributed algorithms which can partition the workload (both data and parameters simultaneously) over a cluster of workers.

In this work, we propose DS-FACTO (Doubly-Separable Factorization Machines), a novel hybrid-parallel \citep{RamSriMatXinYunVis19} stochastic algorithm to scale factorization machines to arbitrarily large workloads. DS-FACTO is based on the NOMAD framework \citep{yun2014nomad} and fully de-centralizes the data as well as model parameters across the workers into mutually exclusive blocks. The key contributions of this work are as follows:
\begin{itemize}
	\item We propose a Hybrid-Parallel algorithm for factorization machines which can partition both the data as well as model parameters simultaneously across the workers.
	\item DS-FACTO follows a fully de-centralized peer-only topology. This avoids the use of parameter servers and associated bottlenecks in a centralized master-slave topology \citep{WatMorFerPie16}.
	\item DS-FACTO is asynchronous and therefore can communicate parameters among the other workers while performing parameter updates.
	\item We present an empirical study that shows such a hybrid-formulation for factorization machines performs competitively as compared to the other existing methods.
\end{itemize}

{\bf Outline}: The rest of the paper is organized as follows. Section \ref{sec:related_work} studies the related work and Section \ref{sec:fm_background} provides some background for factorization machines. In Section \ref{sec:dsfacto}, we introduce DS-FACTO and describe our hybrid-parallelization approach. Section \ref{sec:expmts} is devoted to empirical study comparing DS-FACTO to some standard baselines and studying its scaling behavior. Finally, Section \ref{sec:conclusion} concludes the paper.

\section{Related Work}
\label{sec:related_work}

{\bf Context-aware recommender systems}: There has been lot of work in using factorization machines as the basis to build recommender systems which take into account the user context for its feature representation. Fast context-aware recommendations with factorization machines \citep{rendle2011fast} discusses feature parameterizations that can incorporate diverse types of context information, and also provide numerical pre-computation techniques to optimize the factorization machine model faster. Gradient Boosting factorization machines \citep{cheng2014gradient} borrows ideas from gradient-boosting to select only a subset of the pairwise feature interactions to provide more accurate context representations.

{\bf Click Through Rate (CTR) prediction}: Field-aware factorization machines (FFM) \citep{juan2016field} make use of a set of latent vector for each feature which is dependent on the context of the features with which the pairwise interactions are computed. This is in contrast with vanilla factorization machines where each features is always provided a single latent vector. Empirically, they observe that FFMs achieve better predictive performance for CTR prediction tasks compared to vanilla FMs and polynomial regression models.

\citep{punjabi2018robust} study robustness of factorization machines and design robust counterparts for factorization machines and field-aware factorization machines using robust optimization principles. \citep{guo2016personalized} propose a novel pairwise ranking model using factorization machines which incorporates implicit feedbacks with content information for the task of personalized ranking. \citep{hong2013co} propose Co-Factorization Machines (CoFM), which can deal with multiple aspects of the dataset where each aspect makes use of a separate FM model. CoFM is able to predict user decisions and modeling user interests through content simultaneously. There has also been ample work in extending factorization machines using advances in neural networks. Neural Factorization machines \citep{he2017neural} proposes a variant of FM to apply non-linear activation functions on the pairwise feature interactions. Likewise, Attentional Factorization Machines \citep{xiao2017attentional} is a method to improve vanilla FM models by learning the importance of each feature interaction from data using a neural attention network. DeepFM \citep{guo2017deepfm} proposes a new neural network architecture for factorization machines that involves training a deep component and an FM component jointly. In a completely different vein, \citep{blondel2015convex} proposes a convex formulation of factorization machines based on the nuclear norm and propose an efficient two-block coordinate descent algorithm to optimize the model. \citep{blondel2016higher} discusses how higher-order feature interactions can be used to build more general factorization machine models. \citep{rendle2013scaling} discusses how factorization machines can be used on relational data and scaled to large datasets.

{\bf Scalability}:
There is very limited work in the direction of developing distributed algorithms for factorization machines. LibFM \citep{Ren12}, is one of the most popular implementations of FM that is based on the original paper \citep{Ren10}, however, it is limited to a single-machine. Later, \citep{LiLiuSmoWan16} proposed DiFacto, which is a popular distributed algorithm for Factorization Machines based on the parameter server framework \citep{LiAndParSmoAhm14}. Parameter Server uses a network of workers and servers to partition the data among the workers (and optionally, the model among the servers) and makes use of the message passing interface (MPI) as the communication paradigm. Di-Facto also proposes strategies to adaptively penalize the model parameters based on the frequency of the observed feature values. As a result, DiFacto is able to handle larger workloads than LibFM and also achieves a good predictive performance. \citep{zhong2016scaling} is another work in this direction which also uses parameter server to provide a distributed algorithm for factorization machines - the key difference being that it uses Hadoop as the distributed framework instead of MPI. Finally, \citep{sun2014parallel} also uses Map-Reduce to parallelize factorization machines.



\section{Background and Preliminaries}
\label{sec:fm_background}
The main goal of predictive modeling is to estimate a function $f: \RR^{D} \rightarrow Y$ which can take as input a real valued feature vector in $D$ dimensions and produce a corresponding output. We call such a function $f$ the score function. The output set $Y$ can take values depending on the task at hand. For example, in the case of regression $Y \in \RR$, while for classification tasks, $Y$ takes a positive/negative label such as $\{+1, -1\}$. In supervised settings, it is also assumed that there is a training dataset consisting of $N$ examples and their corresponding labels $\rbr{\xb_{i}, y_{i}}_{i=1, \ldots, N}$, where each $\xb_{i}$ is a $D$-dimensional feature vector.

\subsection{Polynomial Regression}
In this work we are concerned with score functions which can compute second-order feature interactions\footnote{The techniques described in this paper also apply to models that compute higher-order feature interactions, however, for simplicity purposes we assume this simplistic setting.} (also known as pairwise features) in the model. One simple way to accomplish this is by using the {\it Polynomial Regression} model which computes the following score function,
\begin{align}
\label{eq:model_eqn_pr}
f\rbr{\xb_{i}} &= w_{0} + \sum_{j=1}^{D} w_{j} \; x_{ij} + \sum_{j=1}^{D}  \sum_{j'=j+1}^{D} w_{jj'} \; x_{ij} \; x_{ij'}
\end{align}
where, $\xb_{i} \in \RR^{D}$ is an example from the dataset $\Xb \in \RR^{N \times D}$, and the parameters of the model are
$w_{0} \in \RR, \quad \wb \in \RR^{D}, \quad \Wb \in \RR^{D \times D}$. $w_{j}$ denotes $j$-th dimension of $\wb$ and $w_{jj'}$ denotes $\rbr{i,j}$-th entry of $\Wb$.

The model equation of polynomial regression in (\ref{eq:model_eqn_pr}) has a few drawbacks. Real-world datasets are often heavily sparse, which means that all pairwise feature values are unlikely to be observed. Therefore, there is not enough information in the data to be modeled by using a weight for every pairwise feature. Such a parameterization reduces the predictive performance. Moreover, the weight matrix for pairwise features occupies $\Ocal \rbr{D^{2}}$ storage which can be a limitation when the number of dimensions are high.

\subsection{Factorization Machines (FM)}
Factorization machines propose a different way to parameterize pairwise interaction between features to overcome the limitations of polynomial regression. FM aims to learn a latent embedding for every feature such that the pairwise interaction between any two features can be parameterized using the dot product of the corresponding latent embeddings. As a result of this parameterization, FM models work well even when the dataset is extremely sparse, since they only rely on first-order feature values being observed in the data. The score function for factorization machines is computed as,
\begin{align}
\label{eq:fm_model}
f\rbr{\xb_{i}} &= w_{0} + \sum_{j=1}^{D} w_{j} \; x_{ij} + \sum_{j=1}^{D}  \sum_{j'=j+1}^{D} \inner{\vb_{j}}{\vb_{j'}} \; x_{ij} \; x_{ij'}
\end{align}
where, the model parameters are $w_{0} \in \RR, \quad \wb \in \RR^{D}, \quad \Vb \in \RR^{D \times K}$. $\vb_{j} \in \RR^{K}$ denotes the $j$-th dimension of $\Vb$ (latent embedding for the $j$-th feature).

Naive computation of the score function in FM seems to require $\Ocal\rbr{K D^{2}}$. However, using a simple rewrite described below \citep{Ren10}, the score function can be computed in $\Ocal\rbr{K D}$.
\begin{align}
\nonumber
\sum_{j=1}^{D} \sum_{j'=j+1}^{D} \inner{\vb_{j}}{\vb_{j'}} x_{ij} \; x_{ij'} 
\nonumber
&= \frac{1}{2} \sum_{j=1}^{D} \sum_{j'=1}^{D} \inner{\vb_{j}}{\vb_{j'}} x_{ij} \; x_{ij'} - \frac{1}{2} \sum_{j=1}^{D} \inner{\vb_{j}}{\vb_{j}} x_{ij} \; x_{ij} \\
\nonumber
&= \frac{1}{2} \sum_{j=1}^{D} \sum_{j'=1}^{D} \sum_{k=1}^{K} v_{jk} \; v_{j'k} \; x_{ij} \; x_{ij'} - \frac{1}{2} \sum_{j=1}^{D} \sum_{k=1}^{K} v_{jk} \; v_{jk} \; x_{ij} \; x_{ij} \\
\nonumber
&= \frac{1}{2} \sum_{k=1}^{K} \cbr{\rbr{ \sum_{j=1}^{D} v_{jk} x_{ij}} \rbr{ \sum_{j'=1}^{D} v_{j'k} x_{ij'}} - \sum_{j=1}^{D} v_{jk}^{2} x_{ij}^{2}} \\
\label{eq:fm_model_lastterm}
&= \frac{1}{2} \sum_{k=1}^{K} \cbr{\rbr{ {\color{blue} \sum_{d=1}^{D} v_{dk} x_{id}}}^{2} - \sum_{j=1}^{D} v_{jk}^{2} x_{ij}^{2}}
\end{align}
In the sums over $j$, only nnz $\rbr{x_{j}}$ have to be summed up. Plugging this rewrite (\ref{eq:fm_model_lastterm}) into the original model equation (\ref{eq:fm_model}), we obtain its simplified form,
\begin{align}
\label{eq:fm_model2}
f\rbr{\xb_{i}} &= w_{0} + \sum_{j=1}^{D} w_{j} \; x_{ij} + \frac{1}{2} \sum_{k=1}^{K} \cbr{\rbr{ {\color{blue} \sum_{d=1}^{D} v_{dk} x_{id}}}^{2} - \sum_{j=1}^{D} v_{jk}^{2} x_{ij}^{2}}
\end{align}

The complete normalized objective function for Factorization Machine model can now be written as,
\begin{align}
\label{eq:fm_obj}
\Lcal \rbr{\wb, \Vb} &= \quad \frac{1}{N} \sum_{i=1}^{N} l\rbr{f\rbr{\xb_{i}}, y_{i}} + \frac{\lambda_{w}}{2} \rbr{\norm{\wb}_{2}^2} + \frac{\lambda_{v}}{2}  \rbr{\norm{\Vb}_{2}^2}
\end{align}
where,
\begin{itemize}
\item $f\rbr{\xb_{i}}$ is given by (\ref{eq:fm_model2})
\item $\lambda_{w}$ and $\lambda_{v}$ are used to regularize the parameters $\wb$ and $\Vb$
\item $l\rbr{\cdot}$ is an appropriate loss function depending on the task at hand (e.g. cross-entropy for binary classification, squared loss for regression).
\end{itemize}

\begin{table*}[h]
  \renewcommand{\arraystretch}{1}
  \scalebox{0.8}{
\begin{tabular}{| p{0.38\linewidth} | p{0.8\linewidth} |} 
\hline
Symbol & Definition \\ \hline \hline
$N$ & number of observations \\ 
$D$ & number of dimensions \\ 
$K$ & number of latent factors \\  \hline
$\Xb = \cbr{\xb_{1}, \ldots, \xb_{N}}$, $\quad$ $\xb_{i} \in \RR^{D}$ & observations (data points) \\ 
$\yb = \cbr{y_{1}, \ldots, y_{N}}$ & observed labels for the observation $\xb_i$. For regression $y_i \in \RR$. For classification $y_i \in \{+1, -1\}$ \\ 
$\wb \in \RR^{D}$, $\Vb \in \RR^{D \times K}$ & parameters of the model \\ 
$G = \cbr{g_{1}, \ldots, g_{N}}$, $\quad$ $A \in \RR^{N \times K}$ & auxiliary variables used in computing the parameter updates \\ 
\hline
$\lambda_{w}, \lambda_{v}$ & regularization hyper-parameters for $\wb$ and $\Vb$ respectively \\
$\eta$ & learning rate hyper-parameter \\ \hline
\end{tabular}}
\caption{Notations for Factorization Machines}
\label{table:notations}
\end{table*}

{\bf Optimization}: The objective function in (\ref{eq:fm_obj}) can be optimized by any gradient based procedure such as gradient descent. Taking the derivatives of $\Lcal \rbr{\wb, \Vb}$ with respect to the parameters $\wb$ and $\Vb$ we obtain the following updates for gradient descent,

\begin{align}
\label{eq:updates_w0}
w_{0}^{t+1} & \leftarrow w_{0}^{t} - \eta \cdot N \\
\nonumber
w_{j}^{t+1} & \leftarrow w_{j}^{t} - \eta \sum_{i=1}^{N} \nabla_{w_{j}} l_{i} \rbr{\wb, \Vb} + \lambda_{w} \; w_{j}^{t} \\
\nonumber
&= w_{j}^{t} - \eta \sum_{i=1}^{N} {\color{blue}G_{i}^{t}} \cdot \nabla_{w_{j}} f \rbr{\xb_{i}} + \lambda_{w} \; w_{j}^{t} \\
\label{eq:updates_w}
&= w_{j}^{t} - \eta \sum_{i=1}^{N} {\color{blue}G_{i}^{t}} \cdot x_{ij} + \lambda_{w} \; w_{j}^{t}
\end{align}

\begin{align}
\nonumber
v_{jk}^{t+1} & \leftarrow v_{jk}^{t} - \eta \sum_{i=1}^{N} \nabla_{v_{jk}} l_{i} \rbr{\wb, \Vb} + \lambda_{v} \; v_{jk}^{t} \\
\nonumber
&= v_{jk}^{t} - \eta \sum_{i=1}^{N} {\color{blue}G_{i}^{t}} \cdot \nabla_{v_{jk}} f \rbr{\xb_{i}} + \lambda_{v} \; v_{jk}^{t} \\
\label{eq:updates_v}
&= v_{jk}^{t} - \eta \sum_{i=1}^{N} {\color{blue}G_{i}^{t}} \cdot \cbr{x_{ij} \rbr{{\color{blue} \sum_{d=1}^{D} v_{dk}^{t} \cdot x_{id}}} - v_{jk}^{t} \; x_{ij}^{2}} + \lambda_{v} \; v_{jk}^{t}
\end{align}

where, the multiplier {\color{blue}$G_{i}^{t}$} involves computing the score function using the parameter values of $\wb$ and $\Vb$ at the $t$-th iteration as follows,
\begin{align}
{\color{blue}G_{i}^{t}} &=  
\begin{cases}
f \rbr{x_{i}} - y_{i}, \quad \quad \text{if squared loss (regression)} \\
\frac{-y_{i}}{1 + \exp \rbr{y_{i} \cdot f_{i} \rbr{x_{i}}}} , \quad \text{if logistic loss (classification)}
\end{cases}
\end{align}

The term ${\color{blue} \sum_{d=1}^{D} v_{dk} x_{id}}$ requires synchronization across all $D$ dimensions and can be pre-computed. Also, in practice, we only need to sum over the non-zero entries per dimension nnz $\rbr{x_{i}}$. We will denote this synchronization term succinctly as $a_{ik}$,
\begin{align}
\label{eq:compute_a_jk}
{\color{blue}a_{ik}} &= \sum_{d=1}^{D} v_{dk}^{t} \cdot x_{id}
\end{align}

\section{Doubly-Separable Factorization Machines (DS-FACTO)}
\label{sec:dsfacto}

In this section, we describe our proposed distributed optimization algorithm for factorization machines DS-FACTO, which is based on double-separability of functions.
We begin by first studying the parameter updates in factorization machines more closely.

\subsection{Stochastic Optimization}
Based on the updates described in (\ref{eq:updates_w}) and (\ref{eq:updates_v}),  one can take stochastic gradients across $\sum_{i=1}^{N}$ and obtain update rules as follows,
\begin{align}
w_{0}^{t+1} &\leftarrow w_{0}^{t} - \eta \cdot 1 \\
\label{eq:dsfacto_sgdupdatew}
w_{j}^{t+1} &\leftarrow w_{j}^{t} - \eta \; {\color{blue}G_{i}^{t}} \cdot x_{ij} + \lambda_{w} \; w_{j}^{t} \\
\label{eq:dsfacto_sgdupdatev}
v_{jk}^{t+1} &\leftarrow v_{jk}^{t} - \eta \; {\color{blue}G_{i}^{t}} \cdot \cbr{x_{ij} {\color{blue} \; a_{ik}} - v_{jk}^{t} \; x_{ij}^{2}} + \lambda_{v} \; v_{jk}^{t}
\end{align}

The above equations show that updates to $w_{j}$, $V_{jk}$ require accessing only the $j$-th dimension of the $i$-th example, except the terms {\color{blue}$G_{i}^{t}$} and {\color{blue}$a_{ik}$} which involve a summation over all dimensions $j=1, \ldots, D$, and therefore require bulk synchronization at iteration $t$. 

\subsection{Distributing the computation in FM updates}
\begin{figure}[H]
  \centering
  \begin{subfigure}[t]{0.45\textwidth}
    \begin{tikzpicture}[scale=0.4]
      \fill[white!40!white] (0,0) rectangle (5,6);
      \fill[green!90!black] (0,1) rectangle (0.5,2);
      \fill[green!90!black] (1,1) rectangle (1.5,2);
      \draw[black] (0,0) rectangle (5,6);
      \node at (-0.5, 3) {$\Xb$};
      \fill[white!40!white] (0,7) rectangle (5,8);
      \fill[red!40!white] (0,7) rectangle (0.5,8);
      \fill[red!40!white] (1,7) rectangle (1.5,8);
      \draw[black] (0,7) rectangle (5,8);
      \node at (-0.5, 7.5) {$\wb$};
      \fill[white!40!white] (0,9) rectangle (5,12);
      \draw[black] (0,9) rectangle (5,12);
      \node at (-0.5, 10.5) {$\Vb$};
      \fill[white!40!white] (6,0) rectangle (7,6);
      \fill[green!90!black] (6,1) rectangle (7,2);      
      \draw[black] (6,0) rectangle (7,6);
      \node at (7.5, 3) {$G$};
    \end{tikzpicture}
    \caption{$w_{j}$ update}
  \end{subfigure}
  \; \; \; \;
  \begin{subfigure}[t]{0.45\textwidth}
    \begin{tikzpicture}[scale=0.4]
      \fill[white!40!white] (0,0) rectangle (5,6);
      \fill[green!90!black] (0,1) rectangle (0.5,2);
      \fill[green!90!black] (1,1) rectangle (1.5,2);
      \draw[black] (0,0) rectangle (5,6);
      \node at (-0.5, 3) {$\Xb$};
      \fill[white!40!white] (0,7) rectangle (5,8);
      \draw[black] (0,7) rectangle (5,8);
      \node at (-0.5, 7.5) {$\wb$};
      \fill[white!40!white] (0,9) rectangle (5,12);
      \fill[red!40!white] (0,10) rectangle (0.5,11);
      \fill[red!40!white] (1,10) rectangle (1.5,11);      
      \draw[black] (0,9) rectangle (5,12);
      \node at (-0.5, 10.5) {$\Vb$};
      \fill[white!40!white] (6,0) rectangle (7,6);
      \fill[green!90!black] (6,1) rectangle (7,2);      
      \draw[black] (6,0) rectangle (7,6);
      \node at (7.5, 3) {$G$};
      \fill[white!40!white] (8,0) rectangle (11,6);
      \fill[green!90!black] (8,1) rectangle (8.5,2); 
      \fill[green!90!black] (9,1) rectangle (9.5,2);            
      \draw[black] (8,0) rectangle (11,6);
      \node at (11.5, 3) {$A$};
    \end{tikzpicture}  
    \caption{$v_{jk}$ update}
  \end{subfigure}
  \caption{Access pattern of parameters while updating $w_{j}$ and $v_{jk}$. Green indicates the
    variable or data point being read, while Red indicates it being updated. Updating $w_{j}$ requires computing {\color{blue}$G_{i}$} and likewise updating $v_{jk}$ requires computing
    {\color{blue}$a_{ik}$}.}
  \label{fig:accesspattern_w_v_updates}
\end{figure}
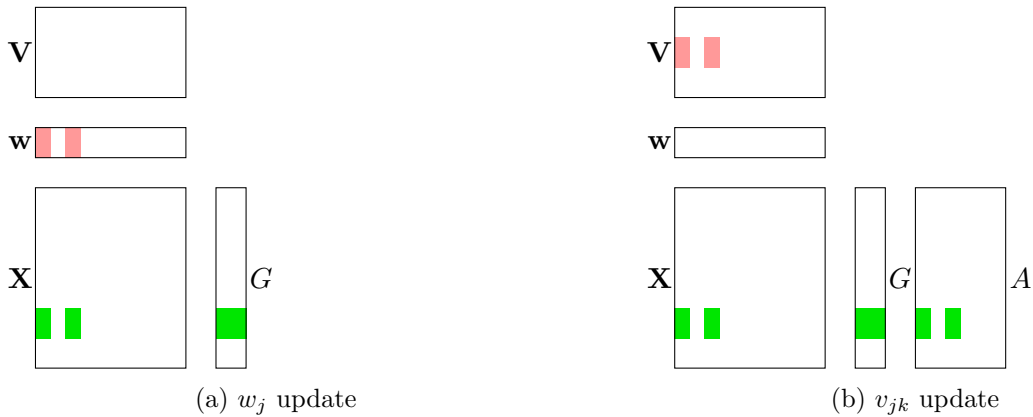

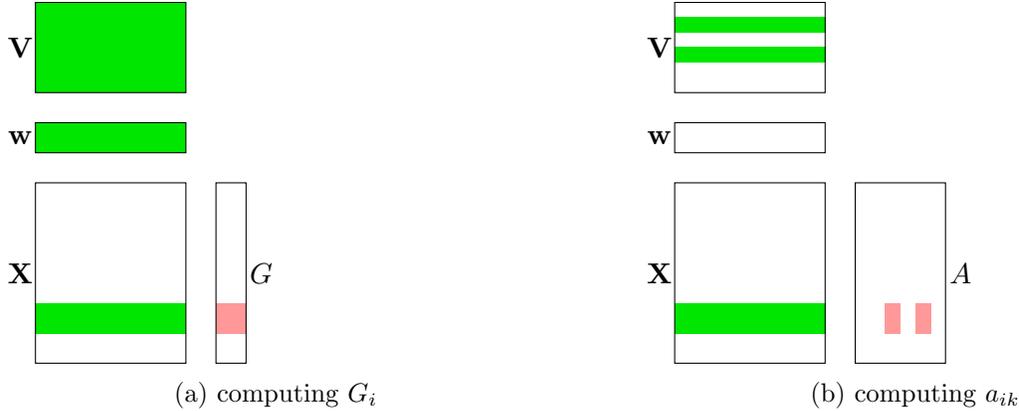
\begin{figure}
  \centering
  \begin{subfigure}[t]{0.45\textwidth}
    \begin{tikzpicture}[scale=0.4]
      \fill[white!40!white] (0,0) rectangle (5,6);
      \fill[green!90!black] (0,1) rectangle (5,2);      
      \draw[black] (0,0) rectangle (5,6);
      \node at (-0.5, 3) {$\Xb$};
      \fill[white!40!white] (0,7) rectangle (5,8);
      \fill[green!90!black] (0,7) rectangle (5,8);
      \draw[black] (0,7) rectangle (5,8);
      \node at (-0.5, 7.5) {$\wb$};
      \fill[white!40!white] (0,9) rectangle (5,12);
      \fill[green!90!black] (0,9) rectangle (5,12);      
      \draw[black] (0,9) rectangle (5,12);
      \node at (-0.5, 10.5) {$\Vb$};
      \fill[white!40!white] (6,0) rectangle (7,6);
      \fill[red!40!white] (6,1) rectangle (7,2);      
      \draw[black] (6,0) rectangle (7,6);
      \node at (7.5, 3) {$G$};
    \end{tikzpicture}
    \caption{computing $G_{i}$}
  \end{subfigure}
  \; \; \; \;
  \begin{subfigure}[t]{0.45\textwidth}
    \begin{tikzpicture}[scale=0.4]
      \fill[white!40!white] (0,0) rectangle (5,6);
      \fill[green!90!black] (0,1) rectangle (5,2);            
      \draw[black] (0,0) rectangle (5,6);
      \node at (-0.5, 3) {$\Xb$};
      \fill[white!40!white] (0,7) rectangle (5,8);
      \draw[black] (0,7) rectangle (5,8);
      \node at (-0.5, 7.5) {$\wb$};
      \fill[white!40!white] (0,9) rectangle (5,12);
      \fill[green!90!black] (0,10) rectangle (5,10.5);
      \fill[green!90!black] (0,11) rectangle (5,11.5);      
      \draw[black] (0,9) rectangle (5,12);
      \node at (-0.5, 10.5) {$\Vb$};
      \fill[white!40!white] (6,0) rectangle (9,6);
      \fill[red!40!white] (7,1) rectangle (7.5,2);            
      \fill[red!40!white] (8,1) rectangle (8.5,2);            
      \draw[black] (6,0) rectangle (9,6);
      \node at (9.5, 3) {$A$};
    \end{tikzpicture}  
    \caption{computing $a_{ik}$}
  \end{subfigure}
  \caption{Access pattern of parameters while computing {\color{blue}$G$} and {\color{blue}$A$}. Green indicates the
    variable or data point being read, while Red indicates it being updated. Observe that computing both {\color{blue}$G$} and {\color{blue}$A$} requires accessing all the dimensions $j=1, \ldots, D$. This is 
    the main synchronization bottleneck.}
  \label{fig:accesspattern_g_a_computation}
\end{figure}

The synchronization terms {\color{blue}$G_{i}^{t}$} and {\color{blue}$a_{i}$}  are the main bottleneck in developing distributed algorithms for factorization machines that are model parallel.
In this section, we study the access patterns of data $\Xb$ and parameters  $\wb$, $\Vb$ during the stochastic gradient descent (SGD) updates.
Figures \ref{fig:accesspattern_w_v_updates} and \ref{fig:accesspattern_g_a_computation} provide a visual illustration. Updating $w_{j}$ and $v_{jk}$ depends on computing $G_{i}$ and $a_{ik}$ respectively (see Figure \ref{fig:accesspattern_w_v_updates}), which unfortunately require synchronization across all the dimensions $j=1, \ldots, D$ (see Figure \ref{fig:accesspattern_g_a_computation}).

{\bf Handling the synchronization terms $G$ and $A$}: In a distributed machine learning
system, there are two popular ways to synchronize parameters,
\begin{itemize}
	\item In a centralized distributed framework following a Map-Reduce paradigm (e.g. parameter server), it is common to perform a {\it Reduce} step where all the workers transmit their copies of the parameters to the server which combines them and transmits them back to each worker.
	\item Another popular way to synchronize using the {\it All-Reduce} paradigm, where all workers transmit their parameter copies to each other.
\end{itemize}
Both of the approaches described above perform {\it Bulk Synchronization} which essentially means they make use of a {\it barrier step} where every worker waits for all other workers to finish their execution. Performing bulk synchronization at every iteration to compute $G$ and $A$ is a huge computational bottleneck.

To resolve this, we propose a different paradigm for synchronization termed as {\it incremental synchronization} \citep{RamSriMatXinYunVis19} which avoids bulk synchronization altogether. The key idea behind incremental synchronization is simple - instead of computing the exact summation or dot product for the synchronization step, we propose computing it incrementally using partial sums. This can be done easily when the workers are arranged in ring topology and follow DSGD style communication (synchronous) or NOMAD style communication (asynchronous).

{\bf Handling the staleness in computing synchronization terms $G$ and $A$}: Since the stochastic updates modify the parameter values of $w_{j}$ and $v_{jk}$ on each worker, the older values of $G_{i}$ and $a_{ik}$ will no longer be up-to-date. This causes some staleness which can slow down convergence significantly. To resolve this, we re-compute $G$ and $A$ after the update step, running an additional set of inner-epochs over all examples $N$ and dimensions $D$. We observed that this re-computation is very important for such a hybrid-parallel scheme to converge correctly.


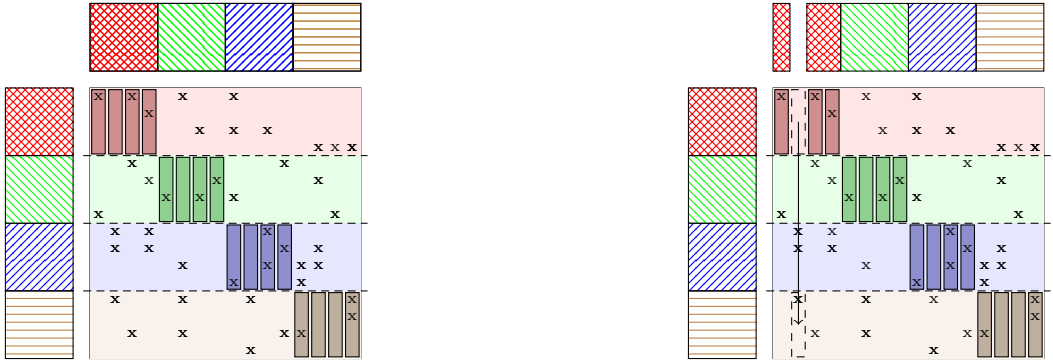
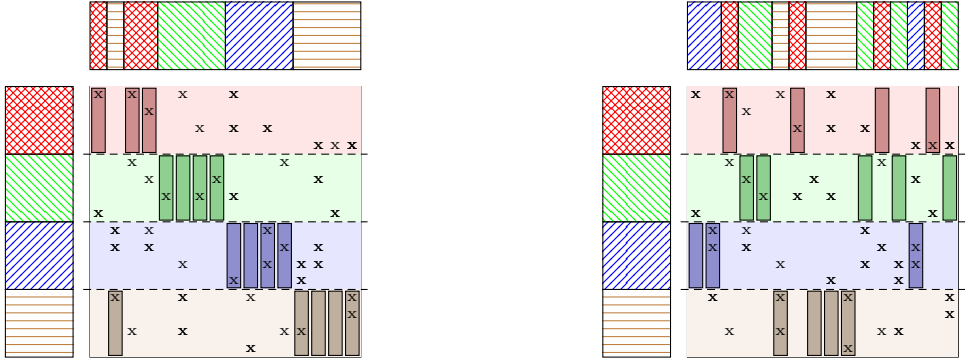
\begin{figure}[!htpb]
  \begin{subfigure}[t]{0.45\textwidth}
    \centering
    \begin{tikzpicture}[scale=0.45]
      
    \draw (0,0) rectangle (\gridwidth * \nodenum, \gridwidth * \nodenum);
  
    \fill [red!10] (0, 3 * \gridwidth) 
    rectangle (\gridwidth * \nodenum, 4 * \gridwidth);

    \fill [green!10] (0, 2 * \gridwidth) 
    rectangle (\gridwidth * \nodenum, 3 * \gridwidth);

    \fill [blue!10] (0, 1 * \gridwidth) 
    rectangle (\gridwidth * \nodenum, 2 * \gridwidth);

    \fill [brown!10] (0,0) 
    rectangle (\gridwidth * \nodenum, 1 * \gridwidth);

    \foreach \y in {1,2,3}
    {
      \draw[densely dashed] 
      (-\gridwidth * 0.1, \gridwidth * \y)
      --
      (\gridwidth * \nodenum + \gridwidth * 0.1,  \gridwidth * \y);
    }
    
    \drawrowpart

    \drawcolpart{0}{3}{\ptrna}{\ptcla}
    \drawcolpart{4}{7}{\ptrnb}{\ptclb}
    \drawcolpart{8}{11}{\ptrnc}{\ptclc}
    \drawcolpart{12}{15}{\ptrnd}{\ptcld}

    \dtpt{5}{1} \dtpt{0}{15} \dtpt{13}{12} \dtpt{13}{6} \dtpt{14}{8}
    \dtpt{0}{8} \dtpt{7}{10} \dtpt{15}{2} \dtpt{10}{5} \dtpt{8}{13}
    \dtpt{5}{3} \dtpt{12}{5} \dtpt{10}{7} \dtpt{11}{6} \dtpt{15}{3}
    \dtpt{6}{9} \dtpt{13}{5} \dtpt{1}{3} \dtpt{8}{15} \dtpt{1}{6}
    \dtpt{1}{7} \dtpt{8}{9} \dtpt{12}{4} \dtpt{12}{1} \dtpt{13}{10}
    \dtpt{9}{0} \dtpt{8}{4} \dtpt{15}{12} \dtpt{3}{6} \dtpt{10}{13}
    \dtpt{6}{13} \dtpt{2}{11} \dtpt{5}{5} \dtpt{2}{1} \dtpt{3}{7}
    \dtpt{9}{3} \dtpt{11}{1} \dtpt{3}{14} \dtpt{11}{11} \dtpt{5}{15}

    \drawcolpart{0}{3}{\ptrna}{\ptcla}
    \drawcolpart{4}{7}{\ptrnb}{\ptclb}
    \drawcolpart{8}{11}{\ptrnc}{\ptclc}
    \drawcolpart{12}{15}{\ptrnd}{\ptcld}

    \asnbox{0}{0}{red} \asnbox{1}{0}{red} \asnbox{2}{0}{red} \asnbox{3}{0}{red}
    \asnbox{4}{1}{green} \asnbox{5}{1}{green} \asnbox{6}{1}{green} \asnbox{7}{1}{green}
    \asnbox{8}{2}{blue} \asnbox{9}{2}{blue} \asnbox{10}{2}{blue} \asnbox{11}{2}{blue}
    \asnbox{12}{3}{brown} \asnbox{13}{3}{brown} \asnbox{14}{3}{brown} \asnbox{15}{3}{brown}
      
    \dtpt{5}{1} \dtpt{0}{15} \dtpt{13}{12} \dtpt{13}{6} \dtpt{14}{8}
    \dtpt{0}{8} \dtpt{7}{10} \dtpt{15}{2} \dtpt{10}{5} \dtpt{8}{13}
    \dtpt{5}{3} \dtpt{12}{5} \dtpt{10}{7} \dtpt{11}{6} \dtpt{15}{3}
    \dtpt{6}{9} \dtpt{13}{5} \dtpt{1}{3} \dtpt{8}{15} \dtpt{1}{6}
    \dtpt{1}{7} \dtpt{8}{9} \dtpt{12}{4} \dtpt{12}{1} \dtpt{13}{10}
    \dtpt{9}{0} \dtpt{8}{4} \dtpt{15}{12} \dtpt{3}{6} \dtpt{10}{13}
    \dtpt{6}{13} \dtpt{2}{11} \dtpt{5}{5} \dtpt{2}{1} \dtpt{3}{7}
    \dtpt{9}{3} \dtpt{11}{1} \dtpt{3}{14} \dtpt{11}{11} \dtpt{5}{15}

    \dtpt{2}{15} \dtpt{3}{10} \dtpt{4}{9} \dtpt{15}{12} \dtpt{14}{12}
      
    \end{tikzpicture}
    \caption{Initial assignment of parameters $\cbr{\wb, \Vb}$ and $\Xb$. Each worker works
      only on the diagonal active area in the beginning.}
  \end{subfigure}
  \quad
  \begin{subfigure}[t]{0.45\textwidth}
    \centering
    \begin{tikzpicture}[scale=0.45]

      \boxlayout
      
      \drawrowpart

      \drawcolpart{0}{0}{\ptrna}{\ptcla}
      \drawcolpart{2}{3}{\ptrna}{\ptcla}
      \drawcolpart{4}{7}{\ptrnb}{\ptclb}
      \drawcolpart{8}{11}{\ptrnc}{\ptclc}
      \drawcolpart{12}{15}{\ptrnd}{\ptcld}

      \asnbox{0}{0}{red} \asnbox{2}{0}{red} \asnbox{3}{0}{red}
      \asnbox{4}{1}{green} \asnbox{5}{1}{green} \asnbox{6}{1}{green} \asnbox{7}{1}{green}
      \asnbox{8}{2}{blue} \asnbox{9}{2}{blue} \asnbox{10}{2}{blue} \asnbox{11}{2}{blue}
      \asnbox{12}{3}{brown} \asnbox{13}{3}{brown} \asnbox{14}{3}{brown} \asnbox{15}{3}{brown}
      
      \dasnbox{1}{0} \dasnbox{1}{3}
      \mvarr{1}{0}{1}{3}

      \dtpt{5}{1} \dtpt{0}{15} \dtpt{13}{12} \dtpt{13}{6} \dtpt{14}{8}
      \dtpt{0}{8} \dtpt{7}{10} \dtpt{15}{2} \dtpt{10}{5} \dtpt{8}{13}
      \dtpt{5}{3} \dtpt{12}{5} \dtpt{10}{7} \dtpt{11}{6} \dtpt{15}{3}
      \dtpt{6}{9} \dtpt{13}{5} \dtpt{1}{3} \dtpt{8}{15} \dtpt{1}{6}
      \dtpt{1}{7} \dtpt{8}{9} \dtpt{12}{4} \dtpt{12}{1} \dtpt{13}{10}
      \dtpt{9}{0} \dtpt{8}{4} \dtpt{15}{12} \dtpt{3}{6} \dtpt{10}{13}
      \dtpt{6}{13} \dtpt{2}{11} \dtpt{5}{5} \dtpt{2}{1} \dtpt{3}{7}
      \dtpt{9}{3} \dtpt{11}{1} \dtpt{3}{14} \dtpt{11}{11} \dtpt{5}{15}
      \dtpt{2}{15} \dtpt{3}{10} \dtpt{4}{9} \dtpt{15}{12} \dtpt{14}{12}

    \end{tikzpicture}
    \caption{After a worker finishes processing column $j$, it sends
      the corresponding parameter set $\cbr{w_{j}, \vb_{j}}$ to another worker.
      Here, $\cbr{w_{2}, \vb_{2}}$ is sent from worker $1$ to $4$.  }
  \end{subfigure}
  \quad \quad
  \begin{subfigure}[t]{0.45\textwidth}
    \centering
    \begin{tikzpicture}[scale=0.45]

      \boxlayout
    
      \drawrowpart

      \drawcolpart{0}{0}{\ptrna}{\ptcla}
      \drawcolpart{1}{1}{\ptrnd}{\ptcld}
      \drawcolpart{2}{3}{\ptrna}{\ptcla}
      \drawcolpart{4}{7}{\ptrnb}{\ptclb}
      \drawcolpart{8}{11}{\ptrnc}{\ptclc}
      \drawcolpart{12}{15}{\ptrnd}{\ptcld}

      \asnbox{0}{0}{red} \asnbox{2}{0}{red} \asnbox{3}{0}{red}
      \asnbox{4}{1}{green} \asnbox{5}{1}{green} \asnbox{6}{1}{green} \asnbox{7}{1}{green}
      \asnbox{8}{2}{blue} \asnbox{9}{2}{blue} \asnbox{10}{2}{blue} \asnbox{11}{2}{blue}
      \asnbox{12}{3}{brown} \asnbox{13}{3}{brown} \asnbox{14}{3}{brown} \asnbox{15}{3}{brown}
      \asnbox{1}{3}{brown}

      \dtpt{5}{1} \dtpt{0}{15} \dtpt{13}{12} \dtpt{13}{6} \dtpt{14}{8}
      \dtpt{0}{8} \dtpt{7}{10} \dtpt{15}{2} \dtpt{10}{5} \dtpt{8}{13}
      \dtpt{5}{3} \dtpt{12}{5} \dtpt{10}{7} \dtpt{11}{6} \dtpt{15}{3}
      \dtpt{6}{9} \dtpt{13}{5} \dtpt{1}{3} \dtpt{8}{15} \dtpt{1}{6}
      \dtpt{1}{7} \dtpt{8}{9} \dtpt{12}{4} \dtpt{12}{1} \dtpt{13}{10}
      \dtpt{9}{0} \dtpt{8}{4} \dtpt{15}{12} \dtpt{3}{6} \dtpt{10}{13}
      \dtpt{6}{13} \dtpt{2}{11} \dtpt{5}{5} \dtpt{2}{1} \dtpt{3}{7}
      \dtpt{9}{3} \dtpt{11}{1} \dtpt{3}{14} \dtpt{11}{11} \dtpt{5}{15}
      \dtpt{2}{15} \dtpt{3}{10} \dtpt{4}{9} \dtpt{15}{12}
      \dtpt{14}{12}

    \end{tikzpicture}
    \caption{Upon receipt, the column is processed by the new worker.
      Here, worker $4$ can now process column $2$ since it owns the
      column.}
  \end{subfigure}
  \quad
  \begin{subfigure}[t]{0.45\textwidth}
    \centering
    \begin{tikzpicture}[scale=0.45]

      \boxlayout

      \drawrowpart

      \asnbox{0}{2}{blue} \asnbox{1}{2}{blue} \asnbox{2}{0}{red} \asnbox{3}{1}{green}
      \asnbox{4}{1}{green} \asnbox{5}{3}{brown} \asnbox{6}{0}{red} \asnbox{7}{3}{brown}
      \asnbox{8}{3}{brown} \asnbox{9}{3}{brown} \asnbox{10}{1}{green} \asnbox{11}{0}{red}
      \asnbox{12}{1}{green} \asnbox{13}{2}{blue} \asnbox{14}{0}{red} \asnbox{15}{1}{green}

      \drawcolpart{0}{1}{\ptrnc}{\ptclc}
      \drawcolpart{2}{2}{\ptrna}{\ptcla}
      \drawcolpart{3}{4}{\ptrnb}{\ptclb}
      \drawcolpart{5}{5}{\ptrnd}{\ptcld}
      \drawcolpart{6}{6}{\ptrna}{\ptcla}
      \drawcolpart{7}{9}{\ptrnd}{\ptcld}
      \drawcolpart{10}{10}{\ptrnb}{\ptclb}
      \drawcolpart{11}{11}{\ptrna}{\ptcla}
      \drawcolpart{12}{12}{\ptrnb}{\ptclb}
      \drawcolpart{13}{13}{\ptrnc}{\ptclc}
      \drawcolpart{14}{14}{\ptrna}{\ptcla}
      \drawcolpart{15}{15}{\ptrnb}{\ptclb}
      
      \dtpt{5}{1} \dtpt{0}{15} \dtpt{13}{12} \dtpt{13}{6} \dtpt{14}{8}
      \dtpt{0}{8} \dtpt{7}{10} \dtpt{15}{2} \dtpt{10}{5} \dtpt{8}{13}
      \dtpt{5}{3} \dtpt{12}{5} \dtpt{10}{7} \dtpt{11}{6} \dtpt{15}{3}
      \dtpt{6}{9} \dtpt{13}{5} \dtpt{1}{3} \dtpt{8}{15} \dtpt{1}{6}
      \dtpt{1}{7} \dtpt{8}{9} \dtpt{12}{4} \dtpt{12}{1} \dtpt{13}{10}
      \dtpt{9}{0} \dtpt{8}{4} \dtpt{15}{12} \dtpt{3}{6} \dtpt{10}{13}
      \dtpt{6}{13} \dtpt{2}{11} \dtpt{5}{5} \dtpt{2}{1} \dtpt{3}{7}
      \dtpt{9}{3} \dtpt{11}{1} \dtpt{3}{14} \dtpt{11}{11} \dtpt{5}{15}

      \dtpt{2}{15} \dtpt{3}{10} \dtpt{4}{9} \dtpt{15}{12} \dtpt{14}{12}

    \end{tikzpicture}
    \caption{During the execution of the algorithm, the ownership of the
      global parameters $\cbr{w_{j}, \vb_{j}}$ changes.}
  \end{subfigure}
  \caption{Illustration of the communication pattern in DS-FACTO algorithm. Parameters $\cbr{w_{j}, \vb_{j}}$ are exchanged in a de-centralized manner across workers without the use of any parameter servers \citep{LiZhoYanLiXia13}.}
  \label{fig:nomad_scheme}
\end{figure}


\subsection{Algorithm}
Algorithm \ref{alg:dsfacto_async} presents a basic outline of DS-FACTO which uses the NOMAD framework \citep{YunYuHsiVisDhi14} for asynchronous communication. The algorithm begins by distributing the data $\Xb$ and parameters $\cbr{\wb, \Vb}$ among $P$ workers as illustrated in Figure (CITE) where the row-blocks represent $\Xb^{(p)}$ and column-blocks represent parameters $\cbr{\wb^{(p)}, \Vb^{(p)}}$ on each local worker respectively. In order to periodically communicate parameter updates across workers, we also maintain $P$ worker queues. The parameters $\cbr{\wb, \Vb}$ are initially distributed uniformly at random across the queues. Each worker can then perform its update in parallel as follows: (1) pops a parameter $(k, \cbr{w_{j}, \vb_{j}})$ out of the queue, (2) updates $w_{j}$ and $v_{jk}$ stochastically using (\ref{eq:dsfacto_sgdupdatew}) and (\ref{eq:dsfacto_sgdupdatev}) respectively, (3) pushes the updated parameter set into the queue of the next worker. Once $D$ rounds of updates have been performed (which is equivalent to saying each worker has updated parameters corresponding to every dimension $j \in \cbr{1, \ldots, D}$), we perform an additional round of communication across the $P$ workers to compute the auxiliary variables $G^{(p)}$ and $A^{(p)}$ using the freshest copy of the parameters. 

\begin{algorithm}[H]
\begin{algorithmic}[1]
	\STATE {$D$: total \# dimensions, \; \; $P$: total \# workers, \; \; $T$: total outer iterations} 
	\STATE{$\cbr{\wb^{(p)}, \Vb^{(p)}}$: parameters per worker, \quad $\cbr{G^{(p)}, A^{(p)}}$: auxiliary variables per worker}
	\STATE{queue[$P$]: array of $P$ worker queues}
	\STATE{Initialize $\wb^{(p)}=0$, $\Vb^{(p)} \sim \Ncal \rbr{0, 0.01}$ \quad {\small \texttt{//Initialize parameters}}}
	\FOR {$j \in \cbr{\wb^{(p)}, \Vb^{(p)}}$}
        		\STATE{Pick $q$ uniformly at random}
        		\STATE{queue[$q$].push($k, \cbr{w_{j}, \vb_{j}}$) \quad \quad \quad \quad \quad {\small \texttt{//Initialize worker queues}}}
        \ENDFOR
	\STATE{{\small \texttt{//Start P workers}}}        
        \FORALL { $p = 1, 2, \ldots, P $ in parallel}
        		\FORALL {$t = 1, 2, \ldots, T$}
			\REPEAT
				\STATE {$(k, \cbr{w_{j}, \vb_{j}}) \leftarrow \text{queue[p].pop()}$}
				\STATE {\text{Update} $w_{j}$ and $v_{jk}$ \text{stochastically using (\ref{eq:dsfacto_sgdupdatew}) and (\ref{eq:dsfacto_sgdupdatev}) }}
				\STATE{Compute index of next queue to push to: $\qhat$}
				\STATE{queue[$\qhat$].push($k, \cbr{w_{j}, \vb_{j}}$)}
			\UNTIL{\# of updates is equal to $D$}
			\REPEAT
				\STATE {$(k, \cbr{w_{j}, \vb_{j}}) \leftarrow \text{queue[p].pop()}$}
				\STATE {Compute $G^{(p)}$ and $A^{(p)}$ using (\ref{eq:fm_model2}) and (\ref{eq:compute_a_jk}) }
			\UNTIL{\# of rounds is equal to $D$}
		\ENDFOR
        \ENDFOR
  \end{algorithmic}
  \caption{DS-FACTO Asynchronous}
  \label{alg:dsfacto_async}
\end{algorithm}

Although the algorithm snippet in Algorithm \ref{alg:dsfacto_async} assumes a restricted setting consisting of $P$ workers, in practice DS-FACTO uses multiple threads on multiple machines. In such a scenario, each worker (thread) first passes around the parameter set across all its threads on its machine. Once this is completed, the parameter set is tossed onto the queue of the first thread on the next machine.

\section{Experiments}
\label{sec:expmts}
In our empirical study, we evaluate DS-FACTO to examine its convergence and scaling behavior. We pick some real-world datasets as shown in Table \ref{table:datasets} for our study.

{\bf Datasets}:
\begin{table}[H]
\center
  \renewcommand{\arraystretch}{1}
  \scalebox{0.95}{
\begin{tabular}{|c|c|c|c|c|}
\hline
Dataset & N & D & K \\ \hline
diabetes & 513 & 8 & 4 \\ 
housing & 303 & 13 & 4 \\ 
ijcnn1 & 49,990 & 22 & 4 \\ 
realsim & 50,616 & 20,958 & 16 \\ \hline
\end{tabular}}
\caption{Dataset Characteristics.}
\label{table:datasets}
\end{table}



\subsection{Convergence and Predictive Performance}
We compare DS-FACTO against libFM \citep{Ren12} which is a widely used library for factorization machines. libFM is a stochastic method which samples the data points stochastically; it however considers all dimensions of the data point while making the parameter updates. DS-FACTO on the other hand is also stochastic in terms of the dimensions; it samples both the data points as well as makes updates only on subsample of the dimensions\footnote{Note that in practice we use incremental gradient descent instead of vanilla stochastic gradient descent in DS-FACTO.}.

\begin{figure}[!htpb]
 \centering
 \begin{tikzpicture}[scale=0.65]
    \begin{axis}[xmode=log, minor tick num=1,
      title={housing dataset},
      xlabel={time (secs)}, ylabel={test RMSE}]

      \addplot[ultra thick, color=blue] table [x index=2, y index=4, header=false, col sep=comma]
      {dsfacto_experiments/serial/dsfacto_housing_1_th1_np1_ppn1_decay1_0.001000_0.100000_0.100000_4_1.log};
      \addlegendentry{DS-FACTO}

      \addplot[ultra thick, color=red] table [x index=2, y index=4, header=false, col sep=comma]
      {dsfacto_experiments/serial/libfm_housing_0.001000_0.100000_0.100000_4_1.log};
      \addlegendentry{LIBFM}

    \end{axis}
  \end{tikzpicture}
   \begin{tikzpicture}[scale=0.65]
    \begin{axis}[xmode=log, minor tick num=1,
      legend style={at={(0.5,0.1)}, anchor= south west},
      title={diabetes dataset},
      xlabel={time (secs)}, ylabel={test accuracy}]

      \addplot[ultra thick, color=blue] table [x index=2, y index=4, header=false, col sep=comma]
      {dsfacto_experiments/serial/dsfacto_diabetes_1_th1_np1_ppn1_decay1_0.010000_0.001000_0.010000_4_1.log};
      \addlegendentry{DS-FACTO}

      \addplot[ultra thick, color=red] table [x index=2, y index=4, header=false, col sep=comma]
      {dsfacto_experiments/serial/libfm_diabetes_0.010000_0.001000_0.010000_4_1.log};
      \addlegendentry{LIBFM}

    \end{axis}
  \end{tikzpicture}
  \begin{tikzpicture}[scale=0.65]
    \begin{axis}[xmode=log, minor tick num=1,
      legend style={at={(0.5,0.1)}, anchor= south west},
      title={ijcnn1 dataset},
      xlabel={time (secs)}, ylabel={test accuracy}]

      \addplot[ultra thick, color=blue] table [x index=2, y index=4, header=false, col sep=comma]
      {dsfacto_experiments/serial/dsfacto_ijcnn1_1_th1_np1_ppn1_decay1_0.001000_0.000100_0.000100_4_1.log};
      \addlegendentry{DS-FACTO}

      \addplot[ultra thick, color=red] table [x index=2, y index=4, header=false, col sep=comma]
      {dsfacto_experiments/serial/libfm_ijcnn1_0.000100_0.000100_0.000100_4_1.log};
      \addlegendentry{LIBFM}

    \end{axis}
  \end{tikzpicture}  
  
%





  \caption{Predictive Performance - Test RMSE (Regression) and Test Accuracy (Classification) of DS-FACTO on {\bf diabetes}, {\bf housing} and {\bf ijcnn1} datasets.}  
 \label{fig:plotRMSE_diabetes_housing}
\end{figure}
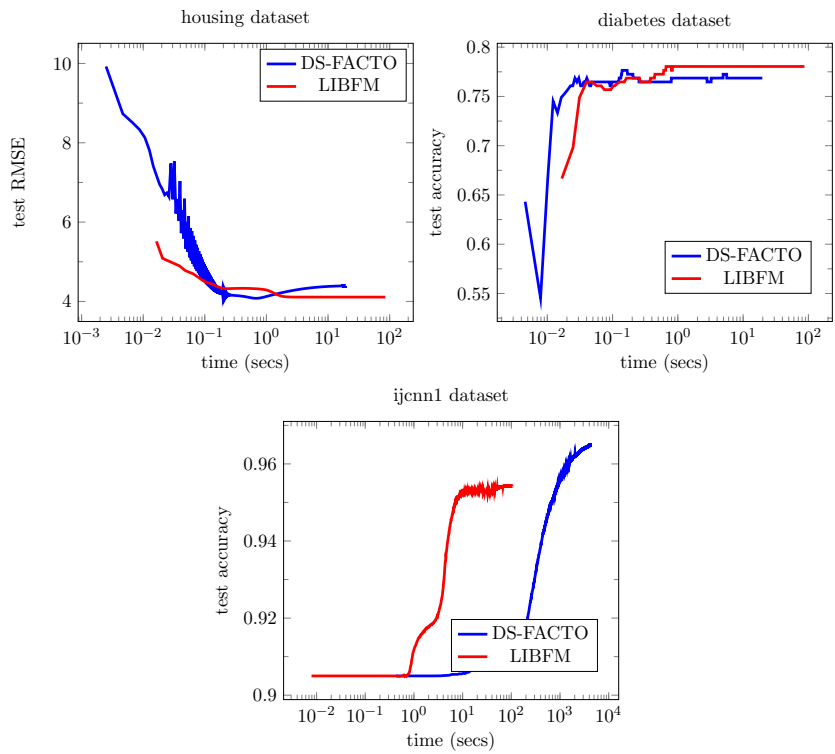

Figures \ref{fig:plottime_diabetes_housing} and \ref{fig:plotRMSE_diabetes_housing} show the convergence behavior and predictive performance of DS-FACTO when compared against libFM. DS-FACTO achieves the similar solution as libFM by making updates just on a subset of dimensions per iteration.

\begin{figure}[!htpb]
 \centering
 \begin{tikzpicture}[scale=0.65]
    \begin{axis}[xmode=log, minor tick num=1,
      title={housing dataset},
      xlabel={time (secs)}, ylabel={objective}]

      \addplot[ultra thick, color=blue] table [x index=2, y index=1, header=false, col sep=comma]
      {dsfacto_experiments/serial/dsfacto_housing_1_th1_np1_ppn1_decay1_0.001000_0.100000_0.100000_4_1.log};
      \addlegendentry{DS-FACTO}

      \addplot[ultra thick, color=red] table [x index=2, y index=1, header=false, col sep=comma]
      {dsfacto_experiments/serial/libfm_housing_0.001000_0.100000_0.100000_4_1.log};
      \addlegendentry{LIBFM}

    \end{axis}
  \end{tikzpicture}
   \begin{tikzpicture}[scale=0.65]
    \begin{axis}[xmode=log, minor tick num=1,
      title={diabetes dataset},
      xlabel={time (secs)}, ylabel={objective}]

      \addplot[ultra thick, color=blue] table [x index=2, y index=1, header=false, col sep=comma]
      {dsfacto_experiments/serial/dsfacto_diabetes_1_th1_np1_ppn1_decay1_0.010000_0.001000_0.010000_4_1.log};
      \addlegendentry{DS-FACTO}

      \addplot[ultra thick, color=red] table [x index=2, y index=1, header=false, col sep=comma]
      {dsfacto_experiments/serial/libfm_diabetes_0.010000_0.001000_0.010000_4_1.log};
      \addlegendentry{LIBFM}

    \end{axis}
  \end{tikzpicture}
  \begin{tikzpicture}[scale=0.65]
    \begin{axis}[xmode=log, minor tick num=1,
      title={ijcnn1 dataset},
      xlabel={time (secs)}, ylabel={objective}]

      
      \addplot[ultra thick, color=blue] table [x index=2, y index=1, header=false, col sep=comma]
      {dsfacto_experiments/serial/dsfacto_ijcnn1_1_th1_np1_ppn1_decay1_0.001000_0.000100_0.000100_4_1.log};

      
      \addplot[ultra thick, color=red] table [x index=2, y index=1, header=false, col sep=comma]
      {dsfacto_experiments/serial/libfm_ijcnn1_0.000100_0.000100_0.000100_4_1.log};

    \end{axis}
  \end{tikzpicture}

%




  \caption{Convergence behavior of DS-FACTO on {\bf diabetes}, {\bf housing} and {\bf ijcnn1} datasets.}
 \label{fig:plottime_diabetes_housing}
\end{figure}
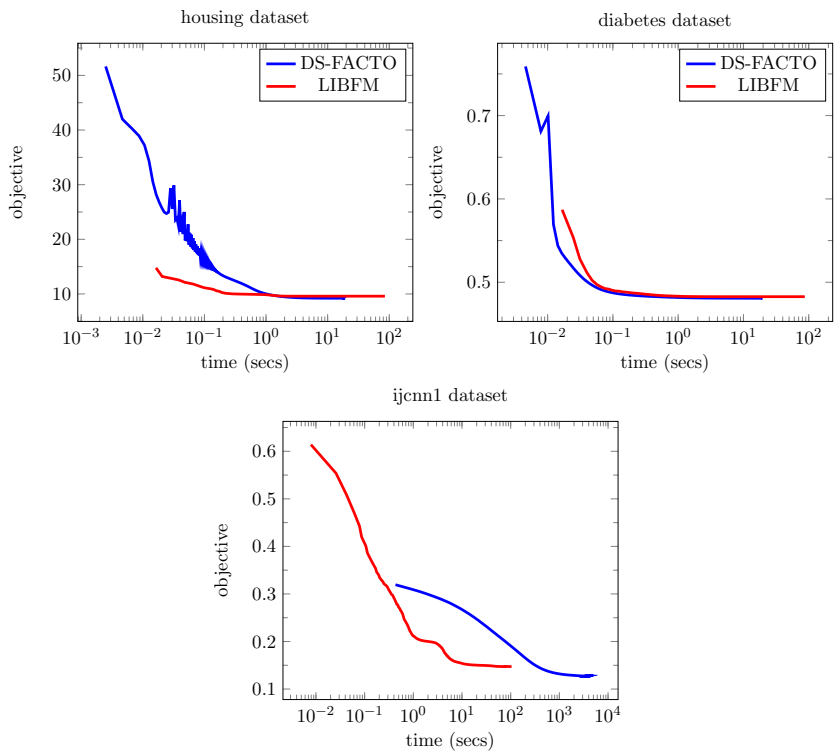

\subsection{Scalability}
\begin{figure}[h]
 \centering
 \begin{tikzpicture}[scale=0.65]
    \begin{axis}[minor tick num=1,
      legend style={at={(0.5,0.01)}, anchor= south west},    
      title={realsim dataset},
      xlabel={time (secs)}, ylabel={speedup}]

      \addplot[ultra thick, dashed, color=black] table [x index=0, y expr=\thisrow{num_p}, header=true]      
      {dsfacto_experiments/parallel/scaling/speedup_th.log};
      \addlegendentry{linear speedup}

      \addplot[ultra thick, mark=*, color=blue] table [x index=0, y expr=\thisrow{time_s}/\thisrow{time_p}, header=true]
      {dsfacto_experiments/parallel/scaling/speedup_th.log};
      \addlegendentry{varying \# of threads}
      
      \addplot[ultra thick, mark=*, color=red] table [x index=0, y expr=\thisrow{time_s}/\thisrow{time_p}, header=true]      
      {dsfacto_experiments/parallel/scaling/speedup_cores.log};
      \addlegendentry{varying \# of cores}
  \end{axis}
  \end{tikzpicture}
  \caption{Scalability of DS-FACTO as \# of threads, cores are varied as 1, 2, 4, 8, 16, 32.}
 \label{fig:scalability}
\end{figure}
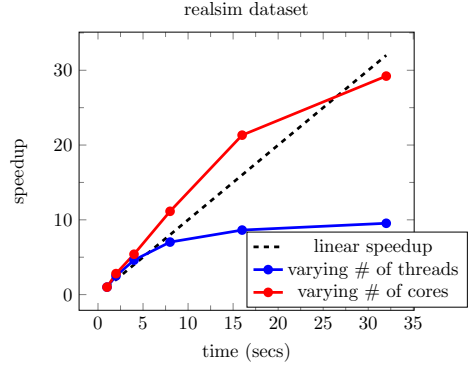

In this sub-section, we present some scalability results, running DS-FACTO in both multi-threaded as well as multi-core architecture. The results of these experiments are shown in Figure \ref{fig:scalability} and the dotted line represents linear speedup. DS-FACTO seems to benefit from multi-core more than multi-threading at this point. One possible reason for this could be that the overheads of adding and removing parameters from the sender and receiver queues while performing the asynchronous communication have high dominating costs.

%
%
%


\section{Conclusion}
\label{sec:conclusion}
In this paper, we presented DS-FACTO, a distributed stochastic optimization algorithm for factorization machines which is hybrid-parallel, i.e. it can partition both data as well as model parameters in a de-centralized manner across workers. In order to circumvent the bulk-synchronization required in computing the gradients for the parameter updates, we make use of local auxiliary variables to maintain partial sums of the synchronization terms and update them in a post-update step. We analyze the behavior of DS-FACTO in terms of convergence and scalability on several real-world datasets. The data partitioning scheme and distributed parameter update strategy used in DS-FACTO is very general and can be easily adapted to scale other variants of factorization machines models such field-aware factorization machines and factorization machines for context-aware recommender systems. We believe these are promising future directions to pursue.

\clearpage
\bibliographystyle{abbrv}
\bibliography{dsfacto}


%

\end{document}